%% file: arxiv_version.tex
\documentclass[runningheads]{llncs}
\usepackage[T1]{fontenc}
\usepackage{graphicx}
\usepackage{amsmath}
\usepackage{amssymb}
\usepackage{multirow}
\usepackage{booktabs}
\usepackage{makecell}
\usepackage{array}
\usepackage{tabularx}
\usepackage{url}
\newcolumntype{L}{>{\raggedright\arraybackslash}X}
\newcolumntype{R}{>{\raggedleft\arraybackslash}X}
\newcolumntype{Y}{>{\centering\arraybackslash}X}

\begin{document}
\title{Verifying Cross-modal Entity Consistency in News using Vision-language Models}


\author{Sahar Tahmasebi\inst{1}\orcidID{0000-0003-4784-7391} \and
David Ernst\inst{3} \and
Eric Müller-Budack\inst{1}\orcidID{0000-0002-6802-1241} \and
Ralph Ewerth\inst{1,2}\orcidID{0000-0003-0918-6297}}
\authorrunning{S. Tahmasebi et al.}
%
\institute{TIB – Leibniz Information Centre for Science and Technology, Hannover, Germany \and
L3S Research Center, Leibniz University Hannover, Hannover, Germany
\and
Leibniz University Hannover, Hannover, Germany
\email{\{sahar.tahmasebi,eric.mueller,ralph.ewerth\}@tib.eu, david.ernst@stud.uni-hannover.de}}


\maketitle 
\begin{abstract}
The web has become a crucial source of information, but it is also used to spread disinformation, often conveyed through multiple modalities like images and text. The identification of inconsistent cross-modal information, in particular entities such as persons, locations, and events, is critical to detect disinformation. Previous works either identify out-of-context disinformation by assessing the consistency of images to the whole document, neglecting relations of individual entities, or focus on generic entities that are not relevant to news. So far, only few approaches have addressed the task of validating entity consistency between images and text in news. However, the potential of large vision-language models (LVLMs) has not been explored yet. In this paper, we propose an LVLM-based framework for verifying Cross-modal Entity Consistency~(LVLM4CEC), to assess whether persons, locations and events in news articles are consistent across both modalities. We suggest effective prompting strategies for LVLMs for entity verification that leverage reference images crawled from web. Moreover, we extend three existing datasets for the task of entity verification in news providing manual ground-truth data. Our results show the potential of LVLMs for automating cross-modal entity verification, showing improved accuracy in identifying persons and events when using evidence images. Moreover, our method outperforms a baseline for location and event verification in documents. The datasets and source code are available on GitHub at \url{https://github.com/TIBHannover/LVLM4CEC}.

\end{abstract}
\begin{figure*}[t]
  \centering
  \includegraphics[width=\textwidth]{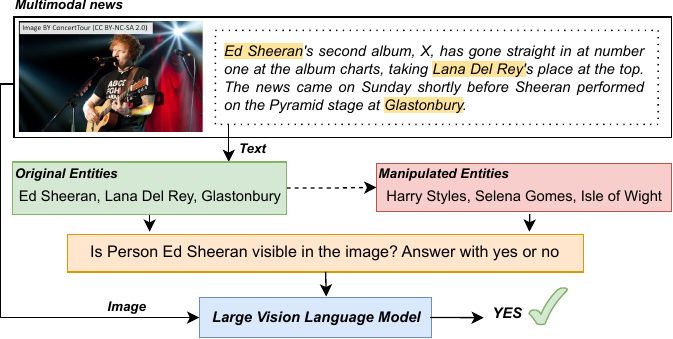}
  \caption{Example of cross-modal entity verification.  Image is replaced with similar one due to license restrictions. Original image is linked on the GitHub.}
  \label{fig:teaser}
\end{figure*}

\section{Introduction}
\input{Sections/Introduction}

\section{Related Work}
\input{Sections/Related_work}

\section{Verifying Cross-Modal Entity Consistency}
\input{Sections/Methodology}

\section{Experimental Setup \& Results}
\input{Sections/Experiments}

\section{Conclusion}
\input{Sections/Conclusion}

\subsubsection{\ackname}
This work was partially funded by the German Federal Ministry of Education
and Research (BMBF, FakeNarratives project, project number 16KIS1517).
We would also like to thank David Ernst~(Leibniz University Hannover) for his valuable input and support.

\subsubsection{\discintname}
The authors have no competing interests to declare that are relevant to the content of this article.
%
%
%
%
\bibliographystyle{splncs04}
%
\bibliography{arxiv_version}
\end{document}

%% file: Sections/Introduction.tex
The web allows media outlets, individuals, etc. to share information and real-time updates
on global events. However, it also enables the spread of misinformation~(false information spread unintentionally), disinformation~(false information spread deliberately), fake news, and other forms of harmful content~\cite{Zubiaga2018}. 
Typically, harmful content is conveyed using different modalities such as images, text, and videos to convey information more efficiently and to attract attention. Additionally, false or harmful information can be created in different ways such as false facts, manipulated photos, or the use of (unaltered) images in a misleading context, also referred to as out-of-context~(OOC) disinformation~(see Figure~\ref{fig:teaser}).
%
%
To cope with the sheer amount of (dis)information shared on the web, automated solution are required. 
One important aspect is to understand cross-modal relations~\cite{bateman2014text,DBLP:gullal} 
to accurately evaluate the overall message of news and help human assessors detecting OOC disinformation~\cite{DBLP:conf/Eric}. 

There are two fields of research related to our work, OOC disinformation detection~\cite{DBLP:Abdelnabi,DBLP:Newsclippings,DBLP:Stefanos,DBLP:sniffer} and works on quantifying general image-text relations~\cite{DBLP:CMI,DBLP:conf/emnlp/KrukLSLJD19,DBLP:Otto_ICMR,DBLP:conf/bmvc/ZhangHK18}. 
Works on OOC disinformation detection~\cite{DBLP:Abdelnabi,DBLP:Newsclippings,DBLP:Stefanos,DBLP:sniffer} focus on the identification of news where unaltered images are used in misleading contexts. 
While these approaches have made substantial progress in recent years, also due to the availability of generative AI models~\cite{DBLP:sniffer}, they predict OOC disinformation based on the consistency of the image to the whole document~(including all entities). However, for many applications it is required to assess the cross-modal consistency of each specific entity~(e.g., persons, locations, and events) individually to understand which entities are potentially misleading. 
On the other hand, approaches that model more general image-text relations such as the cross-modal semantic or objective relations~\cite{DBLP:CMI,DBLP:conf/emnlp/KrukLSLJD19,DBLP:Otto_ICMR,DBLP:conf/bmvc/ZhangHK18} focus on generic entities~(e.g., objects) rather than entities such as persons, locations, and events relevant in news. To date, the only work addressing entity coherence between image and text in news articles is proposed by Müller-Budack et al.~\cite{DBLP:conf/Eric}. 
However, to date, there exist no solution that leverages the capabilities of large vision-language models~(LVLMs), which have demonstrated impressive abilities on a range of tasks including
misinformation detection~\cite{DBLP:chartcheck,DBLP:sniffer,DBLP:journals/sahar}, 
to measure the cross-modal consistency of specific entities between images and text.

In this paper, we propose a pipeline for \textbf{c}ross-modal \textbf{e}ntity \textbf{c}onsistency using \textbf{l}arge \textbf{v}ision-\textbf{l}anguage \textbf{m}odels (\textbf{LVLM4CEC}). The main contributions can be summarized as follows: (1)~We effectively validate the cross-modal consistency of  persons, locations and events in multimodal news articles using LVLMs in zero-shot settings~(Figure~\ref{fig:teaser}). (2)~We propose 
multiple strategies to leverage evidence images for cross-modal entity consistency verification. We present solutions for LVLMs that are limited to a single input image~\cite{DBLP:Instruct-blip,DBLP:llava} and those capable of processing multiple images simultaneously~\cite{DBLP:Mantis,DBLP:DeepSeek}. (3)~
We extend three existing datasets~\cite{DBLP:conf/Eric,DBLP:MMG} for the novel task of cross-modal entity verification and provide manual ground-truth data for two of them. Experiments conducted on these datasets, covering news from different news providers and countries, demonstrate the efficiency of the proposed solution. The results demonstrated impressive zero-shot performance and improved  accuracy for verifying persons and events using evidence images. Additionally, we outperform a baseline~\cite{DBLP:conf/Eric} for document verification based on locations and events.

The remainder of this paper is structured as follows. Section~\ref{sec:related_work}
reviews related work on OOC disinformation detection and image-text relations. Our pipeline for verifying cross-modal entity consistency based on LVLMs is presented in Section~\ref{sec:methodology}. The experimental setup and
results, including the dataset annotation process 
are presented in Section~\ref{sec:experiments}.
Section~\ref{sec:conclusion} concludes the paper and outlines directions for future work.

%% file: Sections/Related_work.tex
\label{sec:related_work}
In this section, we review related work on out-of-context disinformation detection (Section \ref{sec:OOC}) and image-text relations (Section \ref{sec:image-text quantification}) highlighting relevance of existing fields while providing existing research gaps. 

\subsection{Out-of-Context (OOC) Disinformation Detection}
\label{sec:OOC}
OOC disinformation, also referred to as image repurposing or 'cheapfake' detection, 
involves the identification of (unaltered) images in a new, but deceptive, context~\cite{fazio2020out}.
Prior methods have relied on pre-trained models to perform internal checks on given image-text pairs. For instance, Luo et al.~\cite{DBLP:Newsclippings} and Papadopoulos et al.~\cite{DBLP:Stefanos} employed multimodal pre-trained models such as \textit{CLIP}~(Contrastive Language–Image Pre-Training,~\cite{DBLP:CLIP}) and \textit{VisualBERT}~(Visual Bidirectional Encoder Representations from Transformers,~\cite{DBLP:visualBert}) for the classification. 
Meanwhile, Sabir et al.~\cite{DBLP:sabir} and Jaiswal et al.~\cite{DBLP:Jaiswal19} explored the use of external resources, like reference datasets, to conduct external validation. Abdelnabi et al.~\cite{DBLP:Abdelnabi} used both text and image retrieval to find relevant web evidence, then evaluating claim-evidence consistency across textual and visual modalities. More recent models like Sniffer~\cite{DBLP:sniffer}, used instruction tuning for recent LVLMs~\cite{DBLP:Instruct-blip,GPT4-AV} to enhance explainability by generating both judgments and explanations.

The key difference between OOC disinformation detection and our approach lies in the scope of inconsistencies detected. In OOC disinformation detection, the entire text (or all entities) may be inconsistent with the image, or the mismatch might occur at the semantic level. This means that the overall meaning or context conveyed by the image and text does not align. However, in certain applications, it is crucial to pinpoint which specific entities are inconsistent from the image, as this can provide more precise information for tasks like fact-checking or verifying visual content. Our approach focuses on detecting inconsistency at the entity level, where specific elements, such as person, events, or locations, mentioned in the text may not correspond to what is visible in image.

\subsection{Image-Text Relations}
\label{sec:image-text quantification}
Works on OOD disinformation detection assess the consistency between different modalities, e.g., image and text. There has also been a lot of research in the communication science that describe various forms of image-text relations~\cite{bateman2014text,DBLP:gullal}. 
So far, only few works~\cite{DBLP:CMI,DBLP:conf/emnlp/KrukLSLJD19,DBLP:Otto_ICMR,DBLP:conf/bmvc/ZhangHK18} have presented computational models to automatically predict image-text relations such as \textit{objective relations} and \textit{semantic correlation}~\cite{DBLP:CMI,DBLP:Otto_ICMR} that are also highly relevant for news~\cite{DBLP:gullal}. 
%
%
However, these methods focus on rather generic entities and disregard relations of relevant entities for news such as persons, locations, and events. 
To the best of our knowledge, the only approach that focuses on cross-modal entity consistency in news is presented by Müller-Budack et al.~\cite{DBLP:conf/Eric}. They propose an unsupervised approach to quantify the cross-modal consistency persons, locations, and events as well as the overall context. 
%
%
This approach relies on fine-tuned CNNs~\cite{MuellerBudack2018a,MuellerBudack2021,DBLP:faceNet} to measure the cross-modal similarity between the news photo and crawled web images for each entity detected in the news text. However, so far, the capabilities of LVLMs~\cite{DBLP:Instruct-blip,DBLP:Mantis,DBLP:llava,DBLP:DeepSeek} that have shown impressive results for many tasks including, but not limited to, image captioning~\cite{DBLP:corr/abs-2306-09265,DBLP:corr/abs-2306-13549}, chart fact-checking \cite{DBLP:chartcheck} and disinformation detection \cite{DBLP:sniffer,DBLP:journals/sahar}, remain unexplored for cross-modal entity verification.


%% file: Sections/Methodology.tex
\begin{figure}[t]
  \centering
\includegraphics[width=\linewidth]{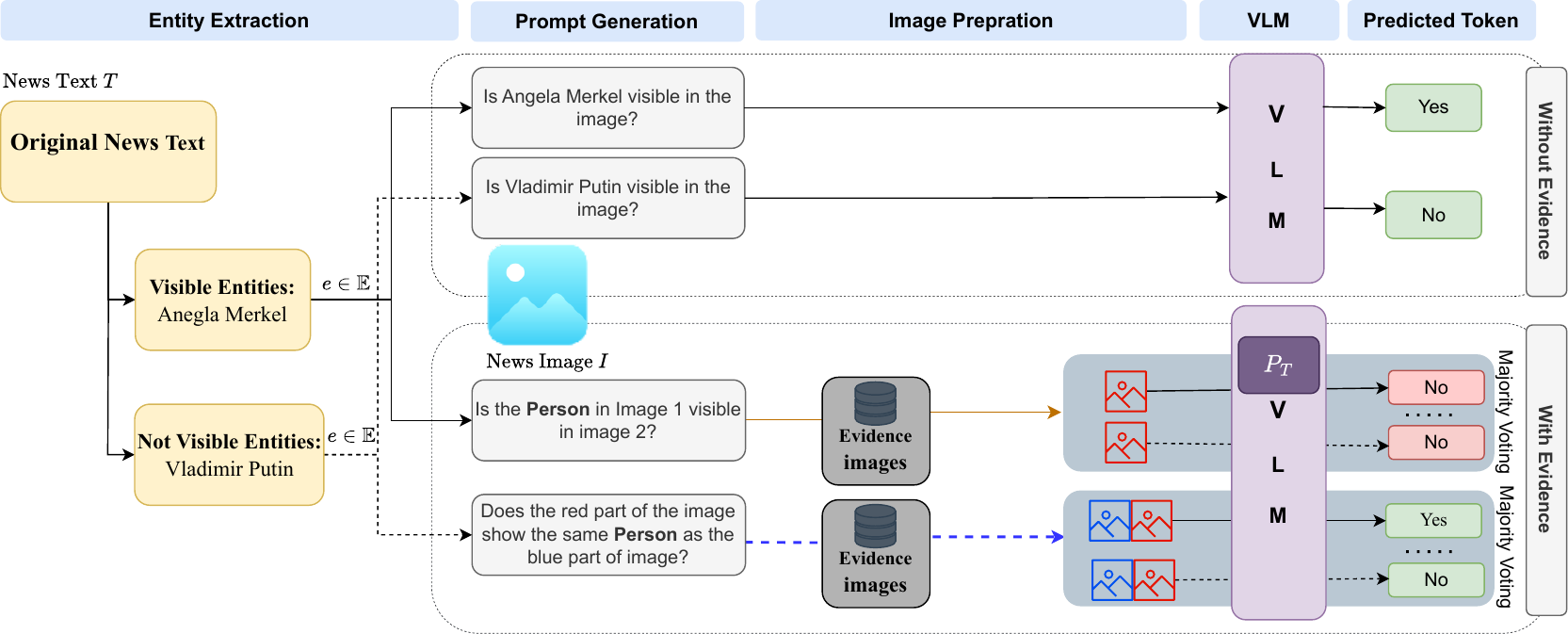}
  \caption{Pipeline for entity consistency verification with~(bottom) and without~(top) using evidence images. The model assesses whether or not an entity~$e \in \mathbb{E}$ is visible, i.e., shares a cross-modal relation, in the news image~$I$. Green indicates valid relations, while red denotes invalid relations.}
  \label{fig:pipeline}
\end{figure}

\label{sec:methodology}
In this section, we propose a
pipeline that exploits LVLMs to assess cross-modal entity consistency in news. An overview is shown in Figure~\ref{fig:pipeline}.
%
%
Given a news article containing both a news image~$I$ and a set of entities~$\mathbb{E}$ extracted from the corresponding news text~$T$, 
the task is to determine whether an entity~$e \in \mathbb{E}$, shares a cross-modal relation~($y = \text{yes}$), i.e., is visible in the news image~$I$, or not~$y = \text{no}$. 
We consider persons, locations, and events as entities.
The pipeline consists of three key steps: entity extraction (Section~\ref{entity_prepration}), prompt generation~(Section \ref{prompt_generation}), and entity verification~(Section \ref{entity_verification}), which explores verification approaches with and without using image evidences crawled from web. 

\subsection{Entity Extraction}
\label{entity_prepration}

To analyze cross-modal consistency, it is essential to first extract entities from the text. Several approaches~\cite{DBLP:ReFinED,DBLP:BLINK} have been introduced for named entity linking~(NEL) in recent years. For comparability to other approaches, we use the approach suggested 
by Müller-Budack et al.~\cite{DBLP:conf/Eric} and  employ \textit{spaCy}~\cite{spacy} and \textit{Wiki\-fier}~\cite{wikifier} to identify persons, locations, and events within a news article. This step captures all necessary entities for subsequent analysis.

\subsection{Prompt Generation}
\label{prompt_generation}
Since our proposed approach operates in a zero-shot setting, selecting an appropriate prompt is crucial, as LVLMs can be sensitive to different prompts.
Furthermore, the questions in the prompts are designed to assess the consistency of each entity~$e \in \mathbb{E}$ depending on its entity type~(person, location, event) and entity verification approach used to integrate evidence images into the prompt~(see Section~\ref{entity_verification}). 
%
%
Based on these parameters, we define different question templates ~(e.g., 
\textit{Is the content of
image consistent with
the <entity type> <entity name>?})
to create a question~$q_e$ for each entity $e\in\mathbb{E}$ detected in the text.
Models are instructed to respond with 'Yes' if the entity is consistent with the image, and 'No' if it is not. In this way, we constrain the model to provide responses within a defined scope which reduces the likelihood of the model generating irrelevant or fabricated information. 
We create various prompt templates for different LVLMs, entity types, and entity verification approaches~(Section~\ref{entity_verification}).
For example, for the location "London Bridge", the prompt \textit{"Is the location of London Bridge shown in the image? Answer with yes or no"} serves as a question when no evidence images are used, 
More examples are shown in Figure~\ref{fig:pipeline}. 
A comparison of prompt templates is conducted in the experiments~(Section~\ref{sec:results}).
\subsection{Entity Verification}
\label{entity_verification}

This step focuses on analyzing the consistency of each entity~$e\in\mathbb{E}$ individually. External evidences images~(e.g., crawled from the Web or knowledge graphs) can improve verification, particularly for less-known entities 
to fully explore this potential, we propose strategies to conduct entity verification both with and without evidence images. 

\subsubsection{Verification without Image Evidence}
\label{sec:Verification_wo_img}
In this strategy, we only provide the news image~$I$ along the question~$q_e$ for each entity~$e\in\mathbb{E}$ detected in the text. 
%
We use prompt-based classification to ensure the model output is robust and we avoid hallucination, by restricting the responses to predefined answers
, ensuring they stay within expected options.  
As each LVLM uses a
tokenizer for text generation with a specific token vocabulary, we categorize tokens into "Yes" and "No" classes, each with all existing variants of yes~(e.g., "Yes", "yes", "YES") and no~(e.g., "no", "No", "NO"). 
in the 
vocabulary. The softmax is applied exclusively on the cumulative sum of the tokens within these two classes. The class with the highest probability determines the generated answer~$y$. 
This approach is suitable for well-known entities that LVLMs can detect without additional information in the form of evidence images. It also only requires one question per entity to generate the answer. However, it may fail for more fine-grained or less-known entities that require evidence images for verification.

\subsubsection{Verification with Image Evidence} 
\label{sec:Verification_w_img}

Contrary to the previous setting, we integrate the news image~$I$ and a maximum of $n$~evidence images for an entity~$e \in \mathbb{E}$ into the question
$q_e$ to assess the cross-model consistency. Evidence images can be collected in various ways. In this work, we follow Müller-Budack et al.~\cite{DBLP:conf/Eric} and use evidence images queried from image search engines~(e.g., Google, Bing) and knowledge graphs~(e.g.,\textit{Wikidata~\cite{wikidata}}) based on the extracted entity from text.  
Choosing a larger number~$n$ of evidence images typically results in a more diverse set of images. This allows for a better entity representation~(e.g., different angles, time frames etc.), particularly for less-known entities, while avoiding noise~(i.e., images that do not depict the queried entity) and potential bias from the search engines. However, it also increases the computational time leading to a trade-off between performance and speed. Moreover, many LVLMs~(e.g.,~\cite{DBLP:Instruct-blip,DBLP:llava}) can only handle a single image as input, while others are capable to process multi-image inputs~(e.g.,~\cite{DBLP:Mantis,DBLP:DeepSeek}). Thus, we present the two following strategies to include evidence images for cross-modal entity verification.
%

For LVLMs limited to a single image input~(e.g.,~\cite{DBLP:Instruct-blip,DBLP:llava}), we propose to create a \textbf{composite image~(\texttt{comp})} that includes the news image~$I$ and \textit{one} evidence image crawled for an entity~$e\in\mathbb{E}$. As shown in Figure~\ref{fig:pipeline}, we add colored borders around these images and ask for their relation in the question~$q_e$, to let the VLM focus on each image individually. To achieve better alignment with the vision encoder's resolution, 
the composite image is created by stacking the images horizontally or vertically, choosing the orientation that yields an aspect ratio closest to 1:1~(quadratic resolution). As the model provides an answer for each of the $n$~evidence images, we use a majority voting for the final decision~$y$ on entity consistency. In case of equal number of votes, we choose the class with the maximum average probability. 

%
For LVLMs that are capable of processing multiple input images simultaneously~(e.g.,~\cite{DBLP:Mantis,DBLP:DeepSeek}), 
we input a series of two images~(\texttt{series}), the news and \textit{one} evidence image, with question~$q_e$ to assess the cross-modal consistency of an entity~$e$. 
Similar to previous section, a majority voting is used for final decision~$y$.

%% file: Sections/Experiments.tex
\label{sec:experiments}
This section provides an overview of the experimental setup (Section~\ref{sec:exp_setup}) and results~(Section~\ref{sec:results}) for cross-modal entity verification. Moreover, Section~\ref{sec:baseline_comparison} conducts a comparison with a baseline on the document verification task. 

\subsection{Experimental Setup}
\label{sec:exp_setup}
In this section, we describe datasets that we have used and annotated specifically for the task of entity verification, 
as well as implementation details.

\subsubsection{Datasets}
\label{sec:dataset}

To cover a wide range of topics, domains, and languages, we use three distinct real-world news datasets, 
\textit{TamperedNews}~\cite{DBLP:conf/Eric}, \textit{News400}~\cite{DBLP:conf/Eric}, and \textit{MMG-NewsPhoto}~\cite{DBLP:MMG} and adapted them for our novel task of quantifying cross-modal entity consistency in news. The dataset statistics are provided in Table~\ref{tab:dataset_overview}. 
\input{tables/dataset_stats}

\textit{TamperedNews} is based on the BreakingNews dataset~\cite{DBLP:breakingNews} and contains news written in English. The dataset contains a set of entities~(persons, locations, events) extracted for each document, along with multiple sets of manipulated entities that are used for a document verification task, i.e., predicting the original set related to the news image. Besides the news image, it also provides evidence images for the entities from various search engines and \textit{Wikidata}. 
\textit{News400} is created in the same way as \textit{TamperedNews} but contains news from three German news websites covering politics, economics, sports, and travel. 
While both datasets provide the original entities extracted from the documents, they do not provide ground-truth annotations for cross-modal consistency of the individual entities, i.e., whether or not the entity mentioned in the text is depicted in the image. Therefore, for both datasets, we manually annotate a random subset called \textbf{TamperedNews-Ent} and \textbf{News400-Ent}. To this end, the cross-modal occurrence of each entity was manually verified by a human annotator. 
The annotator was allowed to use the web for the verification.
%
%
The final dataset contains around 100 samples per entity type for each dataset. The evidence images of the original dataset~\cite{DBLP:conf/Eric} were used for the experiments.

Multimodal geolocation estimation of news photos~(\textbf{MMG-NewsPhoto}) contains 3000 news images  that are manually labeled for their ground-truth geolocation across three spatial resolutions~(city, country, and continent). In contrast to the aforementioned datasets, it does not contain persons and events. 
Since it already provides ground-truth annotations for locations that are visible in the image, we apply the location manipulation techniques proposed by Müller-Budack et al.~\cite{DBLP:conf/Eric} to create negative examples and select a subset of 200 image-entity pairs for our \textbf{MMG-Ent} dataset. The dataset is only used for experiments without evidence images, since no evidence images are provided.

\subsubsection{Metrics}
We use \texttt{Accuracy}, which indicates the proportion of correct answers provided by the model, and \texttt{Unknown Response Rate (URR)}, which quantifies the proportion of answers that are in an unknown or incorrect format, to identify instances where the model fail to generate a valid response.

\subsubsection{Models}
We choose different  LVLMs up to 13B parameters based on their performance on public benchmark datasets like \textit{SeedBench~\cite{DBLP:seedbench}}. For models with single input we choose \textit{BLIP-2}~(Bootstrapping Language-Image Pre-training,~\cite{DBLP:BLIP2}), \textit{InstructBLIP}~\cite{DBLP:Instruct-blip} and \textit{LLaVA 1.5}~(Large Language and Vision Assistant,~\cite{DBLP:llava}), while we use \textit{Mantis}~\cite{DBLP:Mantis} and \textit{DeepSeek}~\cite{DBLP:DeepSeek} to handle multi-image inputs.

\subsubsection{Implementation Details}
\label{sec:implementation}
All experiments were conducted on two NVIDIA A3090, 24-GB GPUs. All models were loaded in 8-bit quantization with a maximum token
length of 256 and the best working prompt according to Section~\ref{sec:prompt_impact}. 

\subsection{Results for Cross-modal Entity Verification}
\label{sec:results}
In this section, we analyze the 
results for cross-modal entity verification using different models, prompts, and verification strategies.
\subsubsection{Prompt Impact}
\label{sec:prompt_impact}

\input{tables/prompts}

We experiment with different prompts~(Section~\ref{prompt_generation}) to evaluate their impact on the model performance. Table~\ref{table:prompts} presents the 
results for different templates when experimenting without evidence images. Results for the other datasets, models and with evidence images allow for the same conclusions and will be added as supplemental material on \textit{GitHub}. 
The results demonstrate that the question design significantly impacts model accuracy, which is expected given that the experiments were conducted in a zero-shot setting. 
\textit{BLIP-2}~\cite{DBLP:BLIP2} hallucinated more than the other models across all templates, reporting a higher \texttt{URR}. Therefore, we excluded it from further experiments. 
Both \textit{InstructBLIP}~\cite{DBLP:Instruct-blip} and \textit{LLaVA}~\cite{DBLP:llava} performed better with \textit{visibility questions} for persons and events entities, while they excelled with \textit{consistency questions} for location entities. 
For each model, we used the best-performing prompt in further experiments.

\subsubsection{Entity Verification without Evidence Images}
\label{sec:results_wo_img}

\input{tables/exp_results}
We evaluate the performance of models on the annotated datasets while considering the entity types person, location and events~(Table~\ref{tab:exp_results}). The results clearly demonstrate that InstructBLIP achieves the best results on \textit{News400-Ent} and \textit{TamperedNews-Ent} across nearly all entity types. In contrast, DeepSeek excels on the \textit{MMG-Ent} dataset, achieving the highest performance at city, country, and continent levels. Another observation is that, when evidence images are excluded, models generally achieve the highest accuracy on location entities, followed by event entities, with person entities showing the lowest performance.

\subsubsection{Entity Verification with Evidence Images}
\label{sec:results_w_img}
To assess the impact of evidence images, we use a maximum of 20~images per entity~(provided by Müller-Budack et al.~\cite{DBLP:conf/Eric}) crawled from Google,  Bing, and Wikidata. 
For models restricted to processing a single image, the image composition strategy~(\texttt{comp}) from Section~\ref{sec:Verification_w_img} is applied. 
For models capable of processing multiple input images simultaneously, we 
provide an image series~(\texttt{series}). 
Results are presented in Table~\ref{tab:exp_results}.

\begin{figure}[t]
  \centering
  \includegraphics[width=0.97\linewidth]{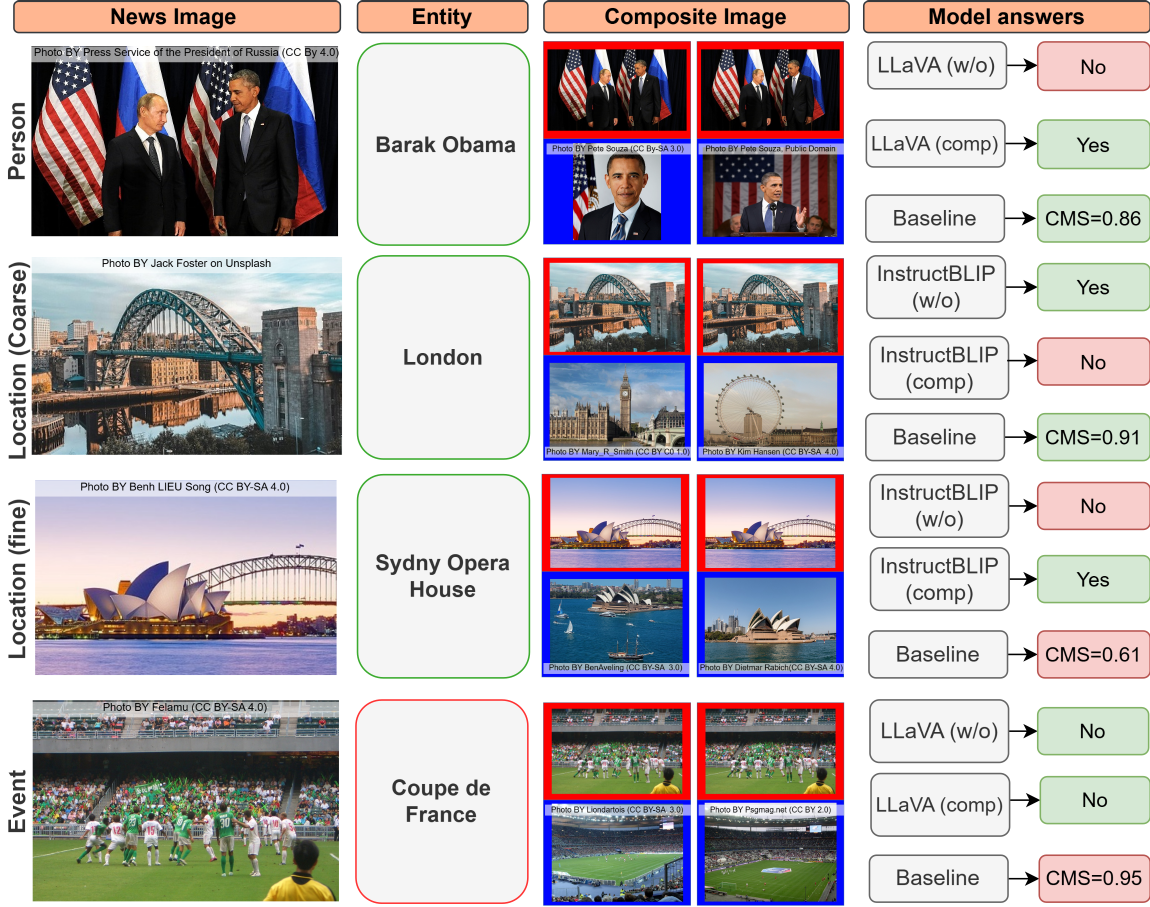}
  \caption{
    Entity verification with and without image evidence across models. 
    Green text box indicates correct predictions; red text indicates incorrect ones. 
    Green borders show visible entities; red borders show invisible ones. 
    As the baseline only outputs a similarity score of Cross-modal Similarities (CMS), 
    we classify CMS values above 0.65 as the 'Yes' class and values below 0.65 as the 'No' class. 
    Images are replaced with similar ones due to license restrictions. 
    Original images are linked on the GitHub.}

  \label{fig:qualitative_exp}
\end{figure}

The results clearly show that evidence images improve the accuracy for person and event entity types. They provide LVLMs with diverse entity representation, encompassing different aspects such as various time frames and perspectives~(see event and person example in Figure \ref{fig:qualitative_exp}).

However, for location entities, the average accuracy either remains unchanged or declines when evidence images are considered. This occurs because evidence images crawled for coarse entities such as cities or countries may not match the place depicted in the news image. Asking for their consistency can lead to wrong predictions as the images are visually dissimilar. For example, evidence image for entity "Hannover" (German or U.S. city), might depict a completely different landmark or building compared to the news image. Also, querying entities such as the city "Liverpool" using Google’s image search engine retrieves images that depict another (more popular) entity, in this case, the football club "Liverpool F.C." rather than the actual entity. The results also indicate that LLaVA outperforms other models on event entities, while the Mantis model achieves the best performance on person entities compared to the other models.

\subsection{Baseline Comparison for Document Verification}
\label{sec:baseline_comparison}
We compare our LVLM-based approach with the method proposed by Müller-Budack et al.~\cite{DBLP:conf/Eric} as baseline. 
Their method evaluates the entity consistency between images and text in news articles using web-sourced evidence images and off-the-shelf CNNs~\cite{MuellerBudack2018a,MuellerBudack2021,DBLP:faceNet} for feature extraction. While it outputs Cross-modal Similarities~(CMS) for each entity, this baseline does not provide a final decision~($y \in \{\text{yes, no}\}$) regarding its consistency.
To enable a fair comparison to our method, we follow the document verification setting proposed by the authors~\cite{DBLP:conf/Eric}. In this setting, the CMS of each entity in the original and \textit{one} tampered entity set are compared. A document is considered to be classified correctly, when an untampered entity achieves the highest CMS. 
%
To evaluate our approach in this setting, we use the probability of generating the token~'yes' as the CMS for each entity.
As the \textit{TamperedNews} and \textit{News400} datasets do not verify whether a relation exists between the image and the original set of entities, we use the manually verified subsets presented in our work. 
The results are shown in Table~\ref{tab:baseline_comparison}. 
\input{tables/baseline_comparison}

The results demonstrate that our approach outperforms the baseline in recognizing entity types locations and events, even in a zero-shot setting~(example shown in Figure \ref{fig:qualitative_exp}). 
While the baseline uses a CNN for event type classification~\cite{MuellerBudack2021} that cannot distinguish more fine-grained event instances, LVLMs can differentiate between such events leading to significantly better results when tampered events of the same event class are used. 
InstructBLIP achieves the best results for location identification, while LLaVA excels in event recognition. 
However, the baseline achieves better results in person identification which is expected since the entities and the retrieved evidence images are very unambiguous, and
neural networks for face recognition, such as \textit{FaceNet}~\cite{DBLP:faceNet} used in this work, already achieve impressive results. 


%% file: tables/dataset_stats.tex
\begin{table}[t]
\centering
\setlength{\tabcolsep}{12pt}
\caption{Dataset statistics including the number documents~($D$), entities~($E$), and entities~($E_\mathtt{vis}$) that have been verified manually to be depicted in the news photo.}
\label{tab:dataset_overview}
\begin{tabularx}
{\linewidth}{X X c c c}
\toprule
\textbf{Dataset} & \textbf{Entity Type}  & \textbf{$D$} & \textbf{$E$} & \textbf{$E_\mathtt{vis}$} \\ 
\midrule
\multirow{4}{*}{\makecell{MMG-NewsPhoto-Ent}} & Cities     & 200  & 123  & 123  \\ 
                                  & Countries  & 200  & 47   & 47   \\ 
                                  & Continents & 200  & 6    & 6    \\ 
\cmidrule(lr){2-5}
                                  & All        & 200  & 176  & 176  \\
\midrule
\multirow{4}{*}{News400-Ent}       & Persons    & 122  & 3498 & 642  \\ 
                                  & Locations  & 67   & 3276 & 926  \\ 
                                  & Events     & 25   & 489  & 137  \\
\cmidrule(lr){2-5}                & All        & 395  & 7263 & 1705 \\                  
\midrule
\multirow{4}{*}{\makecell{TamperedNews-Ent}} & Persons    & 104  & 2940 & 104  \\ 
                                  & Locations  & 100  & 2419 & 123  \\ 
                                  & Events     & 98   & 501  & 66   \\
\cmidrule(lr){2-5}                & All        & 266  & 5860 & 293  \\                  
\bottomrule
\end{tabularx}
\end{table}

%% file: tables/prompts.tex
\begin{table}[t]
\centering
\setlength{\tabcolsep}{4pt}
\caption{Accuracy~(ACC) and unknown response rate~(URR) for different LVLMs and entity types~(ET) using various question templates on TamperedNews-Ent.}
\label{table:prompts}
\begin{tabularx}{\linewidth}{l c YYYYYY }
\hline
\multirow{2}{*}{\textbf{Question Template}} & \multirow{2}{*}{\textbf{ET}} & \multicolumn{2}{c}{\textbf{BLIP-2}} & \multicolumn{2}{c}{\textbf{InstructBLIP}} & \multicolumn{2}{c}{\textbf{LLaVA-1.5}} \\ \cmidrule(lr){3-4} \cmidrule(lr){5-6} \cmidrule(lr){7-8}
 & & \textbf{ACC} & \textbf{URR} & \textbf{ACC} & \textbf{URR} & \textbf{ACC} & \textbf{URR} \\ \hline
 
\multirow{3}{*}{\makecell[l]{Is $<$type$>$ $<$name$>$\\ shown in the image?}} & Person   & 0.39 & 0.06 & \textbf{0.73} & 0.00 & 0.52 & 0.00  \\ 
  & Location & 0.47 & 0.08 & \textbf{0.72} & 0.00 & 0.52 & 0.00  \\ 
  & Event    & 0.48 & 0.11 & 0.59 & 0.00 & 0.76 & 0.00  \\ \hline
  
\multirow{3}{*}{\makecell[l]{Is $<$type$>$ $<$name$>$\\ shown in the image?\\ Answer with yes or no.}} & Person   & 0.33 & 0.03 & 0.65 & 0.01 & \textbf{0.68} & 0.00  \\ 
  & Location & 0.37 & 0.03 & 0.64 & 0.00 & \textbf{0.64} & 0.00  \\ 
  & Event    & 0.63 & 0.11 & 0.83 & 0.00 & 0.70 & 0.00  \\ \hline
  
\multirow{3}{*}{\makecell[l]{Is the content of the \\image consistent with\\ the $<$type$>$ $<$name$>$?}} & Person   & 0.39 & 0.06 & 0.60 & 0.01 & 0.29 & 0.02  \\ 
  & Location & 0.30 & 0.03 & 0.60 & 0.00 & 0.44 & 0.08  \\ 
  & Event    & \textbf{0.70} & 0.04 & \textbf{0.87} & 0.00 & \textbf{0.78} & 0.09  \\ \hline

\multirow{3}{*}{\makecell[l]{Is any $<$type$>$ from\\ the image consistent\\ with $<$name$>$?}} & Person   & \textbf{0.78} & 0.00 & 0.70 & 0.01 & 0.51 & 0.03  \\ 
  & Location & \textbf{0.69} & 0.02 & 0.67 & 0.00 & 0.47 & 0.01  \\ 
  & Event    & 0.20 & 0.00 & 0.63 & 0.00 & 0.67 & 0.22  \\ \hline
\end{tabularx}
\end{table}

%% file: tables/exp_results.tex
\begin{table}[t]

\centering
\caption{Accuracy for entity verification using different models on three datasets for entity types persons~(PER), locations~(LOC), and events~(EVT). For MMG-News various spatial resolution for locations, i.e., city~(LCt), country~(LCo), and continent~(LCn) are considered. Results are shown by image composition strategies: without evidence images (w/o), single image composition~(comp), and multiple image series.}
\label{tab:exp_results}
\footnotesize
\begin{tabularx}{\textwidth}{l c *{3}{>{\centering\arraybackslash}X} *{3}{>{\centering\arraybackslash}X} *{3}{>{\centering\arraybackslash}X}}
\toprule
\textbf{Model}  & \textbf{ICS}  & \multicolumn{3}{c}{\makecell{\textbf{Tampered}\\\textbf{News-Ent}}} & \multicolumn{3}{c}{\makecell{\textbf{News400}\\\textbf{-Ent}}} & \multicolumn{3}{c}{\textbf{MMG-Ent}} \\
\cmidrule(lr){3-5} \cmidrule(lr){6-8} \cmidrule(lr){9-11}
& &  \textbf{PER} & \textbf{LOC} & \textbf{EVT} & \textbf{PER} & \textbf{LOC} & \textbf{EVT} & \textbf{LCt} & \textbf{LCo} & \textbf{LCn} \\
\midrule

\multicolumn{11}{c}{\textbf{Models with Single Image Input}} \\
\midrule

\multirow{2}{*}{\textbf{InstructBLIP}}
 &  w/o & 0.66 & 0.81 & 0.76 & 0.68 & 0.75 & 0.79 & 0.63 & 0.30 & 0.59 \\
 &  comp  & 0.73 & 0.78 & 0.72 & 0.71 & 0.67 & 0.85 & - & - & -\\
\hline

\multirow{2}{*}{\textbf{LLaVA 1.5}}
 & w/o & 0.61 & 0.79 & 0.71 & 0.63 & 0.70 & 0.57 & 0.70 & 0.48 &  0.27 \\
 &  comp &  0.78 & 0.73 & \textbf{0.77} & 0.77 & 0.70 & \textbf{0.85} & - & - & - \\
 
\midrule
\multicolumn{11}{c}{\textbf{Models with Multiple Image Input}} \\
\midrule

\multirow{2}{*}{\textbf{Mantis}}
 & w/o &  0.63 & 0.79 & 0.73 & 0.68 & 0.74 & 0.60 & 0.60 & 0.26 & 0.11 \\
 &  series &  \textbf{0.84} & 0.74 & 0.74  & \textbf{0.80} & 0.70 & 0.72 & - &  - & - \\
\hline

\multirow{3}{*}{\textbf{DeepSeek}}
 & w/o &  0.66 & 0.79 & 0.70 & 0.65 & 0.75 & 0.61 & 0.74 & 0.52 & 0.76 \\
 & series &  0.80 & 0.62 & 0.72 & 0.78 & 0.61 & 0.80 & - & - & -\\
\bottomrule
\end{tabularx}
\end{table}

%% file: tables/baseline_comparison.tex
\begin{table}[t]
\centering
\caption{Average accuracy of InstructBLIP~(InstBLIP~\cite{DBLP:Instruct-blip}) and LLaVA~(\cite{DBLP:llava}) compared to the baseline from Müller-Budack et al.~\cite{DBLP:conf/Eric} on TamperedNews-Ent and News400-Ent for different entity types~(ET). For persons, manipulation strategies~(MS) of same country (PsC), same gender (PsG) and same country and gender (PsCG), for locations, different Great Circle Distance (GCD) range $\text{GCD}_{min}^{max}$, and for events, replacements with events on same parent class are used.}
\label{tab:baseline_comparison}
\setlength{\tabcolsep}{2pt}
\begin{tabularx}{\linewidth}{llYYYYYY}
\toprule
\multirow{2}{*}{\textbf{ET}} & \multirow{2}{*}{\textbf{MS}} & \multicolumn{3}{c}{\textbf{TamperedNews-Ent}} & \multicolumn{3}{c}{\textbf{News400-Ent}} \\
\cmidrule(lr){3-5} \cmidrule(lr){6-8}
& & \textbf{Baseline} & \textbf{InstBlip} & \textbf{LLaVA} & \textbf{Baseline} & \textbf{InstBlip} & \textbf{LLaVA} \\
\midrule
\multirow{4}{*}{PER} 
& Random & \textbf{0.94} & 0.86 & 0.86 & \textbf{0.95} & 0.90 & 0.91 \\
& PsC    & \textbf{0.93} & 0.86 & 0.84 & \textbf{0.92} & 0.81 & 0.83 \\
& PsG    & \textbf{0.94} & 0.84 & 0.81 & \textbf{0.91} & 0.86 & 0.85 \\
& PsCG   & \textbf{0.93} & 0.79 & 0.78 & \textbf{0.92} & 0.86 & 0.79 \\
\midrule
\multirow{4}{*}{LOC} 
& Random & 0.81 & \textbf{0.91} & 0.96 & 0.85 & \textbf{0.96} & 0.91 \\
& $\text{GCD}_{25}^{200}$  & 0.80 & \textbf{0.87} & 0.80 & 0.80 & \textbf{0.90} & 0.87 \\
& $\text{GCD}_{200}^{750}$ & 0.77 & \textbf{0.93} & 0.91 & 0.84 & \textbf{0.93} & 0.91 \\
& $\text{GCD}_{750}^{2500}$ & 0.73 & \textbf{0.91} & 0.89 & 0.74 & \textbf{0.87} & 0.90 \\
\midrule
\multirow{2}{*}{EVT} 
& Random & 0.92 & 0.89 & \textbf{0.97} & \textbf{0.97} & 0.96 & \textbf{0.97} \\
& Same-class & 0.75 & 0.90 & \textbf{0.94} & 0.74 & 0.84 & \textbf{0.92} \\
\bottomrule
\end{tabularx}
\end{table}

%% file: Sections/Conclusion.tex
\label{sec:conclusion}
In this paper, we suggested an approach for verifying Cross-modal Entity Consistency using Large Vision-Language Models (LVLM4CEC). 
We introduce effective zero-shot approaches capable of assessing the consistency of person, location, and event entities in multimodal news with and without evidence images crawled from the web.
Unlike prior works on out-of-context disinformation detection that predict image consistency across the whole document~(or all entities), we focus on analyzing each entity allowing to assess which entities may be misleading in news articles.
For this purpose, we extended three existing datasets and provided TamperedNews-Ent, News400-Ent, and MMG-Ent for the task of entity verification including manual annotations for a reliable evaluation. 
%
The experiments demonstrated the efficiency of the proposed solution. Notably, the performance of LVLMs improved for cross-modal person and event verification when incorporating evidence images. Furthermore, our approach outperformed a baseline for the crucial task of location and event verification in documents. Datasets and source code are publicly available on GitHub.

In the future, we will focus on improving the cross-modal verification of locations by retrieving more representative photos. Since broad location queries~(e.g., continents, countries, cities) often lack specificity, using Google Image Search or a predefined dataset of key global locations could improve results. Furthermore, beyond the entity types focused on in this study—persons, locations, and events—verification of other entities, such as times (e.g., decades, dates, daytime) and organizations, remains an open area for exploration.

%% file: arxiv_version.bbl
\begin{thebibliography}{10}
\providecommand{\url}[1]{\texttt{#1}}
\providecommand{\urlprefix}{URL }
\providecommand{\doi}[1]{https://doi.org/#1}

\bibitem{DBLP:Abdelnabi}
Abdelnabi, S., Hasan, R., Fritz, M.: Open-domain, content-based, multi-modal fact-checking of out-of-context images via online resources. In: {IEEE/CVF} Conference on Computer Vision and Pattern Recognition, {CVPR} 2022, New Orleans, LA, USA, June 18-24, 2022. pp. 14920--14929. {IEEE} (2022). \doi{10.1109/CVPR52688.2022.01452}

\bibitem{DBLP:chartcheck}
Akhtar, M., Subedi, N., Gupta, V., Tahmasebi, S., Cocarascu, O., Simperl, E.: Chartcheck: An evidence-based fact-checking dataset over real-world chart images. arXiv preprint  \textbf{abs/2311.07453} (2023). \doi{10.48550/ARXIV.2311.07453}

\bibitem{DBLP:ReFinED}
Ayoola, T., Tyagi, S., Fisher, J., Christodoulopoulos, C., Pierleoni, A.: Refined: An efficient zero-shot-capable approach to end-to-end entity linking. In: Conference of the North American Chapter of the Association for Computational Linguistics: Human Language Technologies: Industry Track, {NAACL} 2022, Hybrid: Seattle, Washington, {USA} + Online, July 10-15, 2022. pp. 209--220. Association for Computational Linguistics (2022). \doi{10.18653/V1/2022.NAACL-INDUSTRY.24}

\bibitem{bateman2014text}
Bateman, J.: Text and image: A critical introduction to the visual/verbal divide. Routledge (2014)

\bibitem{wikifier}
Brank, J., Leban, G., Grobelnik, M.: Annotating documents with relevant wikipedia concepts. Proceedings of SiKDD  \textbf{472} (2017)

\bibitem{DBLP:gullal}
Cheema, G.S., Hakimov, S., M{\"{u}}ller{-}Budack, E., Otto, C., Bateman, J.A., Ewerth, R.: Understanding image-text relations and news values for multimodal news analysis. Frontiers in Artificial Intelligence  \textbf{6} (2023). \doi{10.3389/FRAI.2023.1125533}

\bibitem{DBLP:Instruct-blip}
Dai, W., Li, J., Li, D., Tiong, A.M.H., Zhao, J., Wang, W., Li, B., Fung, P., Hoi, S.C.H.: Instructblip: Towards general-purpose vision-language models with instruction tuning. In: Annual Conference on Neural Information Processing Systems, NeurIPS 2023, New Orleans, LA, USA, December 10 - 16, 2023 (2023), \url{https://openreview.net/forum?id=vvoWPYqZJA}

\bibitem{fazio2020out}
Fazio, L.: Out-of-context photos are a powerful low-tech form of misinformation. The Conversation  \textbf{14}, ~1 (2020)

\bibitem{DBLP:CMI}
Henning, C.A., Ewerth, R.: Estimating the information gap between textual and visual representations. In: International Conference on Multimedia Retrieval, {ICMR} 2017, Bucharest, Romania, June 6-9, 2017. pp. 14--22. {ACM} (2017). \doi{10.1145/3078971.3078991}

\bibitem{spacy}
Honnibal, M., Montani, I.: {spaCy 2: Natural language understanding with Bloom embeddings, convolutional neural networks and incremental parsing}. To appear  \textbf{7}(1),  411--420 (2017)

\bibitem{DBLP:Jaiswal19}
Jaiswal, A., Wu, Y., AbdAlmageed, W., Masi, I., Natarajan, P.: {AIRD:} adversarial learning framework for image repurposing detection. In: {IEEE} Conference on Computer Vision and Pattern Recognition, {CVPR} 2019, Long Beach, CA, USA, June 16-20, 2019. pp. 11330--11339. Computer Vision Foundation / {IEEE} (2019). \doi{10.1109/CVPR.2019.01159}

\bibitem{DBLP:Mantis}
Jiang, D., He, X., Zeng, H., Wei, C., Ku, M., Liu, Q., Chen, W.: {MANTIS:} interleaved multi-image instruction tuning. arXiv preprint  \textbf{abs/2405.01483} (2024). \doi{10.48550/ARXIV.2405.01483}

\bibitem{DBLP:conf/emnlp/KrukLSLJD19}
Kruk, J., Lubin, J., Sikka, K., Lin, X., Jurafsky, D., Divakaran, A.: Integrating text and image: Determining multimodal document intent in instagram posts. In: Conference on Empirical Methods in Natural Language Processing and the International Joint Conference on Natural Language Processing, {EMNLP-IJCNLP} 2019, Hong Kong, China, November 3-7, 2019. pp. 4621--4631. Association for Computational Linguistics (2019). \doi{10.18653/V1/D19-1469}

\bibitem{DBLP:seedbench}
Li, B., Wang, R., Wang, G., Ge, Y., Ge, Y., Shan, Y.: Seed-bench: Benchmarking multimodal llms with generative comprehension. arXiv preprint  \textbf{abs/2307.16125} (2023). \doi{10.48550/ARXIV.2307.16125}

\bibitem{DBLP:BLIP2}
Li, J., Li, D., Savarese, S., Hoi, S.C.H.: {BLIP-2:} bootstrapping language-image pre-training with frozen image encoders and large language models. In: International Conference on Machine Learning, {ICML} 2023, 23-29 July 2023, Honolulu, Hawaii, {USA}. pp. 19730--19742. {PMLR} (2023), \url{https://proceedings.mlr.press/v202/li23q.html}

\bibitem{DBLP:visualBert}
Li, L.H., Yatskar, M., Yin, D., Hsieh, C., Chang, K.: Visualbert: {A} simple and performant baseline for vision and language. arXiv preprint  \textbf{abs/1908.03557} (2019)

\bibitem{DBLP:llava}
Liu, H., Li, C., Wu, Q., Lee, Y.J.: Visual instruction tuning. In: Annual Conference on Neural Information Processing Systems, NeurIPS 2023, New Orleans, LA, USA, December 10 - 16, 2023 (2023), \url{https://openreview.net/forum?id=w0H2xGHlkw}

\bibitem{DBLP:DeepSeek}
Lu, H., Liu, W., Zhang, B., Wang, B., Dong, K., Liu, B., Sun, J., Ren, T., Li, Z., Yang, H., Sun, Y., Deng, C., Xu, H., Xie, Z., Ruan, C.: Deepseek-vl: Towards real-world vision-language understanding. arXiv preprint  \textbf{abs/2403.05525} (2024). \doi{10.48550/ARXIV.2403.05525}

\bibitem{DBLP:Newsclippings}
Luo, G., Darrell, T., Rohrbach, A.: Newsclippings: Automatic generation of out-of-context multimodal media. In: Conference on Empirical Methods in Natural Language Processing, {EMNLP} 2021, Virtual Event / Punta Cana, Dominican Republic, 7-11 November, 2021. pp. 6801--6817. Association for Computational Linguistics (2021). \doi{10.18653/V1/2021.EMNLP-MAIN.545}

\bibitem{MuellerBudack2018a}
M{\"{u}}ller{-}Budack, E., Pustu{-}Iren, K., Ewerth, R.: Geolocation estimation of photos using a hierarchical model and scene classification. In: European Conference on Computer Vision, {ECCV} 2018, Munich, Germany, September 8-14, 2018. pp. 575--592. Springer (2018). \doi{10.1007/978-3-030-01258-8\_35}

\bibitem{MuellerBudack2021}
M{\"{u}}ller{-}Budack, E., Springstein, M., Hakimov, S., Mrutzek, K., Ewerth, R.: Ontology-driven event type classification in images. In: {IEEE} Winter Conference on Applications of Computer Vision, {WACV} 2021, Virtual Event, January 3-8, 2021. pp. 2927--2937. {IEEE} (2021). \doi{10.1109/WACV48630.2021.00297}

\bibitem{DBLP:conf/Eric}
M{\"{u}}ller{-}Budack, E., Theiner, J., Diering, S., Idahl, M., Ewerth, R.: Multimodal analytics for real-world news using measures of cross-modal entity consistency. In: International Conference on Multimedia Retrieval, {ICMR} 2020, Dublin, Ireland, June 8-11, 2020. pp. 16--25. {ACM} (2020). \doi{10.1145/3372278.3390670}

\bibitem{GPT4-AV}
OpenAI: Gpt-4v(ision) system card (2023), \url{https://cdn.openai.com/papers/GPTV_System_Card.pdf}

\bibitem{DBLP:Otto_ICMR}
Otto, C., Springstein, M., Anand, A., Ewerth, R.: Understanding, categorizing and predicting semantic image-text relations. In: International Conference on Multimedia Retrieval, {ICMR} 2019, Ottawa, ON, Canada, June 10-13, 2019. pp. 168--176. {ACM} (2019). \doi{10.1145/3323873.3325049}

\bibitem{DBLP:Stefanos}
Papadopoulos, S., Koutlis, C., Papadopoulos, S., Petrantonakis, P.: Synthetic misinformers: Generating and combating multimodal misinformation. In: International Workshop on Multimedia {AI} against Disinformation, MAD@ICMR 2023, Thessaloniki, Greece, June 12-15, 2023. pp. 36--44. {ACM} (2023). \doi{10.1145/3592572.3592842}

\bibitem{DBLP:sniffer}
Qi, P., Yan, Z., Hsu, W., Lee, M.: Sniffer: Multimodal large language model for explainable out-of-context misinformation detection. In: {IEEE/CVF} Conference on Computer Vision and Pattern Recognition, {CVPR} 2024, Seattle, WA, USA, June 16-22, 2024. pp. 13052--13062. {IEEE} (2024). \doi{10.1109/CVPR52733.2024.01240}

\bibitem{DBLP:CLIP}
Radford, A., Kim, J.W., Hallacy, C., Ramesh, A., Goh, G., Agarwal, S., Sastry, G., Askell, A., Mishkin, P., Clark, J., Krueger, G., Sutskever, I.: Learning transferable visual models from natural language supervision. In: International Conference on Machine Learning, {ICML} 2021, Virtual Event, 18-24 July, 2021. pp. 8748--8763. {PMLR} (2021), \url{http://proceedings.mlr.press/v139/radford21a.html}

\bibitem{DBLP:breakingNews}
Ramisa, A., Yan, F., Moreno{-}Noguer, F., Mikolajczyk, K.: Breakingnews: Article annotation by image and text processing. IEEE Transactions on Pattern Analysis and Machine Intelligence  \textbf{40}(5),  1072--1085 (2018). \doi{10.1109/TPAMI.2017.2721945}

\bibitem{DBLP:sabir}
Sabir, E., AbdAlmageed, W., Wu, Y., Natarajan, P.: Deep multimodal image-repurposing detection. In: {ACM} Multimedia Conference on Multimedia Conference, {MM} 2018, Seoul, Republic of Korea, October 22-26, 2018. pp. 1337--1345. {ACM} (2018). \doi{10.1145/3240508.3240707}

\bibitem{DBLP:faceNet}
Schroff, F., Kalenichenko, D., Philbin, J.: Facenet: {A} unified embedding for face recognition and clustering. In: {IEEE} Conference on Computer Vision and Pattern Recognition, {CVPR} 2015, Boston, MA, USA, June 7-12, 2015. pp. 815--823. {IEEE} Computer Society (2015). \doi{10.1109/CVPR.2015.7298682}

\bibitem{DBLP:journals/sahar}
Tahmasebi, S., M\"{u}ller-Budack, E., Ewerth, R.: Multimodal misinformation detection using large vision-language models. In: International Conference on Information and Knowledge Management, CIKM 2024, Boise, ID, USA, October 21 – 25 2024. p. 2189–2199. Association for Computing Machinery, New York, NY, USA (2024). \doi{10.1145/3627673.3679826}

\bibitem{DBLP:MMG}
Tahmasebzadeh, G., Hakimov, S., Ewerth, R., M{\"{u}}ller{-}Budack, E.: Multimodal geolocation estimation of news photos. In: European Conference on Information Retrieval, {ECIR} 2023, Dublin, Ireland, April 2-6, 2023, Proceedings, Part {II}. pp. 204--220. Springer (2023). \doi{10.1007/978-3-031-28238-6\_14}

\bibitem{wikidata}
Vrande{\v{c}}i{\'c}, D., Kr{\"o}tzsch, M.: Wikidata: a free collaborative knowledgebase. Communications of the ACM  \textbf{57}(10),  78--85 (2014). \doi{10.1145/2629489}

\bibitem{DBLP:BLINK}
Wu, L., Petroni, F., Josifoski, M., Riedel, S., Zettlemoyer, L.: Scalable zero-shot entity linking with dense entity retrieval. In: Conference on Empirical Methods in Natural Language Processing, {EMNLP} 2020, Online, November 16-20, 2020. pp. 6397--6407. Association for Computational Linguistics (2020). \doi{10.18653/V1/2020.EMNLP-MAIN.519}

\bibitem{DBLP:corr/abs-2306-09265}
Xu, P., Shao, W., Zhang, K., Gao, P., Liu, S., Lei, M., Meng, F., Huang, S., Qiao, Y., Luo, P.: Lvlm-ehub: {A} comprehensive evaluation benchmark for large vision-language models. arXiv preprint  \textbf{abs/2306.09265} (2023). \doi{10.48550/ARXIV.2306.09265}

\bibitem{DBLP:corr/abs-2306-13549}
Yin, S., Fu, C., Zhao, S., Li, K., Sun, X., Xu, T., Chen, E.: A survey on multimodal large language models. arXiv preprint  \textbf{abs/2306.13549} (2023). \doi{10.48550/ARXIV.2306.13549}

\bibitem{DBLP:conf/bmvc/ZhangHK18}
Zhang, M., Hwa, R., Kovashka, A.: Equal but not the same: Understanding the implicit relationship between persuasive images and text. In: British Machine Vision Conference 2018, {BMVC} 2018, Newcastle, UK, September 3-6, 2018. p.~8. {BMVA} Press (2018), \url{http://bmvc2018.org/contents/papers/0228.pdf}

\bibitem{Zubiaga2018}
Zubiaga, A., Aker, A., Bontcheva, K., Liakata, M., Procter, R.: Detection and resolution of rumours in social media: {A} survey. ACM Computing Surveys  \textbf{51}(2),  32:1--32:36 (2018). \doi{10.1145/3161603}

\end{thebibliography}
